\documentclass[sigconf, authorversion]{acmart}
\usepackage{subcaption}
\AtBeginDocument{%
  }

\copyrightyear{2026}
\acmYear{2026}
\setcopyright{cc}
\setcctype{by}
\acmConference[ETRA '26]{2026 Symposium on Eye Tracking Research and Applications}{June 01--04, 2026}{Marrakesh, Morocco}
\acmBooktitle{2026 Symposium on Eye Tracking Research and Applications (ETRA '26), June 01--04, 2026, Marrakesh, Morocco}
\acmDOI{10.1145/3797246.3806217}
\acmISBN{979-8-4007-2519-7/2026/06}

\begin{document}

\title{An Affordable, Wearable Stereo-Eye-Tracking Platform}

\author{Alexander Zimmer}
\email{alexander.zimmer@tum.de}
\orcid{0009-0003-8800-104X}
\affiliation{%
	\institution{Human-Centered Technologies for Learning, Technical University of Munich}
 \city{Munich}
 \country{Germany}
}

\author{Yasmeen Abdrabou}
\email{yasmeen.abdrabou@tum.de}
\orcid{0000-0002-8895-4997}
\affiliation{%
	\institution{Human-Centered Technologies for Learning, Technical University of Munich}
 \city{Munich}
 \country{Germany}
}

\author{Enkelejda Kasneci}
\email{enkelejda.kasneci@tum.de}
\orcid{0000-0003-3146-4484}
\affiliation{%
	\institution{Human-Centered Technologies for Learning, Technical University of Munich}
 \city{Munich}
 \country{Germany}}

\renewcommand{\shortauthors}{Zimmer et al.}

\begin{abstract}
Research on video-based eye-tracking has long explored stereo and glint-based methods, yet existing wearable eye trackers — both commercial and open-source — offer limited flexibility for algorithm development and comparative evaluation. We present an affordable, wearable stereo eye-tracking platform built from off-the-shelf and 3D-printable components that explicitly targets this gap. The system combines four infrared eye cameras, infrared illumination, an optional scene camera, and software support for calibration and synchronized data acquisition. By design, the platform supports multiple eye-tracking paradigms, including stereo, glint-based, and binocular approaches, within a single hardware configuration. Rather than optimizing for end-user robustness, the platform prioritizes modularity and extensibility for research use. This paper focuses on the hardware architecture and calibration pipeline and demonstrates the feasibility of the approach using a prototype implementation. All hardware designs and documentation are made openly available
\footnote{Repo: \url{https://github.com/VIVARefSys/WearableEyeTracker}}
\end{abstract}

\ccsdesc[500]{Hardware~Emerging technologies}
\ccsdesc[500]{Human-centered computing~Human computer interaction (HCI)}
\keywords{Wearable eye tracking, Stereo eye tracking, Research platform}


\maketitle
\section{Introduction}





Video-based eye tracking has become a core sensing modality in human–computer interaction~\cite{Majaranta2014} and psychology~\cite{duchowskiBreadthfirstSurveyEyetracking2002}, enabling detailed analysis of visual attention, perception, and behavior. Over the past decades, a wide range of eye-tracking approaches have been proposed, including appearance-based methods~\cite{zhangEvaluationAppearanceBasedMethods2019,krafkaEyeTrackingEveryone2016}, model-based gaze estimation~\cite{swirskiFullyautomaticTemporalApproach2013,santiniGetGripSlippagerobust2019}, stereo reconstruction~\cite{shihNovelApproach3D2004}, and glint-based techniques that exploit corneal reflections of known light sources~\cite{barsingerhornDevelopmentValidationHighspeed2018}. Despite this methodological diversity, progress in the field is increasingly constrained not by algorithmic ideas alone, but by the limited availability of flexible, wearable hardware platforms that allow researchers to develop, compare, and benchmark these approaches under consistent sensing conditions. A particularly underexplored gap lies in wearable stereo eye tracking that simultaneously supports glint-based illumination. Existing wearable eye tracking devices, both commercial~\cite{kassner2014Pupil} and open-source~\cite{etvr}, are typically optimized for a single tracking paradigm and prioritize robustness and ease of use for end users. As a result, they offer limited access to raw, synchronized multi-camera data and restrict experimentation with alternative camera geometries, illumination setups, and calibration strategies. This makes it difficult to systematically study algorithmic trade-offs, reproduce prior work, or develop hybrid approaches that combine multiple paradigms.


Addressing this gap is important because hardware design choices fundamentally shape what eye-tracking algorithms can be developed and evaluated. Without access to adaptable sensing platforms, researchers are forced either to modify closed systems or to build ad hoc prototypes that are difficult to reproduce and compare. A dedicated research-oriented platform can instead enable principled experimentation, fair benchmarking, and more transparent evaluation of eye-tracking methods, particularly in wearable contexts where geometric constraints are critical.

In this work, we present a novel hardware setup for wearable video-based stereo eye tracking designed explicitly for algorithm development and comparative research, rather than end-user deployment. The proposed system integrates four infrared eye cameras, infrared illumination suitable for glint-based methods, and an optional scene camera into a modular, head-mounted configuration. The platform supports multiple eye-tracking paradigms, including stereo, glint-based, and binocular model-based approaches, within a single hardware design. We describe the complete hardware architecture, component selection, mechanical design, and a calibration pipeline for synchronized multi-camera acquisition.

The key novelty of our work lies in providing, to the best of our knowledge, the first openly available wearable eye-tracking platform that combines stereo cameras and infrared spot illumination in a unified, modular design.




\section{Related work}

\subsection{Video-Based Eye Tracking}

Video-based eye tracking encompasses a wide range of approaches with different sensing assumptions, hardware requirements, and calibration demands \cite{duchowski2017Eye,hansen2010In}. On one end, appearance-based methods estimate gaze directly from eye images, often using learning-based mappings and monocular cameras, resulting in low hardware complexity but increased sensitivity to illumination changes and head pose \cite{swirski2012Robust}. In contrast, model-based approaches explicitly represent the 3D geometry of the eye and typically achieve higher accuracy and interpretability, at the cost of precise calibration and controlled imaging conditions~\cite{hansen2010In,guestrin2006General}. Building on these models, stereo and binocular methods leverage multiple eye cameras to reconstruct ocular geometry in 3D, reducing estimation ambiguity but requiring tightly synchronized multi-camera setups that are uncommon in wearable systems~\cite{wang20133D}. On the other side, glint-based techniques introduce active infrared illumination to generate corneal reflections that serve as stable reference points, improving robustness to head movement but adding further hardware and synchronization complexity~\cite{guestrin2006General}.

\subsection{Wearable Eye-Tracking Systems}
Wearable eye trackers are typically implemented as head-mounted systems to support gaze estimation under natural head and body movement~\cite{duchowski2017Eye}. Commercial devices are largely optimized for robustness and ease of use, often abstracting hardware details and limiting access to raw sensor data, which constrains experimentation with alternative sensing configurations and algorithms~\cite{hansen2010In}. From a research perspective, wearable systems must balance competing constraints such as size, weight, power consumption, and user comfort, frequently resulting in simplified designs with monocular cameras and fixed illumination~\cite{kassner2014Pupil}. This reflects a broader trade-off between robustness and flexibility: while tightly integrated systems favor stable operation, they limit reconfigurability, leaving few wearable platforms suitable for algorithm development and comparative evaluation across eye-tracking paradigms~\cite{kassner2014Pupil, santiniPuReRobustPupil2018}.

\subsection{Open-Source and Research-Oriented Eye-Tracking Platforms}
Several open-source eye-tracking platforms have improved accessibility and reproducibility in the field, most notably the Pupil framework, which provides an extensible wearable system based on monocular infrared eye cameras~\cite{kassner2014Pupil}, but ultimately relies on proprietary hardware. Other projects focus purely on webcam eye-tracking~\cite{optikey,gazetracking} or mono camera setups~\cite{eivaziInconspicuousModularHeadmounted2018}. Hosp et al.\cite{hospRemoteEyeOpensourceHighspeed2020a} describe an open-source high-speed remote eye-tracker, which enables high performance at an affordable price point, but is not suitable for wearable applications. The EyeTrackVR\cite{etvr} project aims to provide eye tracking for several VR headsets. It uses binocular mono-cameras and a glint-based tracking approach.

Existing open-source platforms are typically built around fixed sensing configurations and offer limited support for more complex setups, such as stereo eye cameras or alternative illumination geometries. As a result, openly available wearable systems that combine stereo vision with glint-based illumination in a unified and reproducible design remain largely absent, limiting systematic comparison of eye-tracking paradigms and the development of hybrid methods.

This paper addresses this research gap by introducing a modular, wearable eye-tracking hardware platform that supports multiple tracking paradigms within a single, reproducible configuration, with a focus on flexibility and algorithmic experimentation rather than end-user optimization.






\section{Hardware Implementation}

\begin{figure}[t!]
    \centering
    \begin{subfigure}[t]{\linewidth}
        \centering
        \includegraphics[height=1.5in]{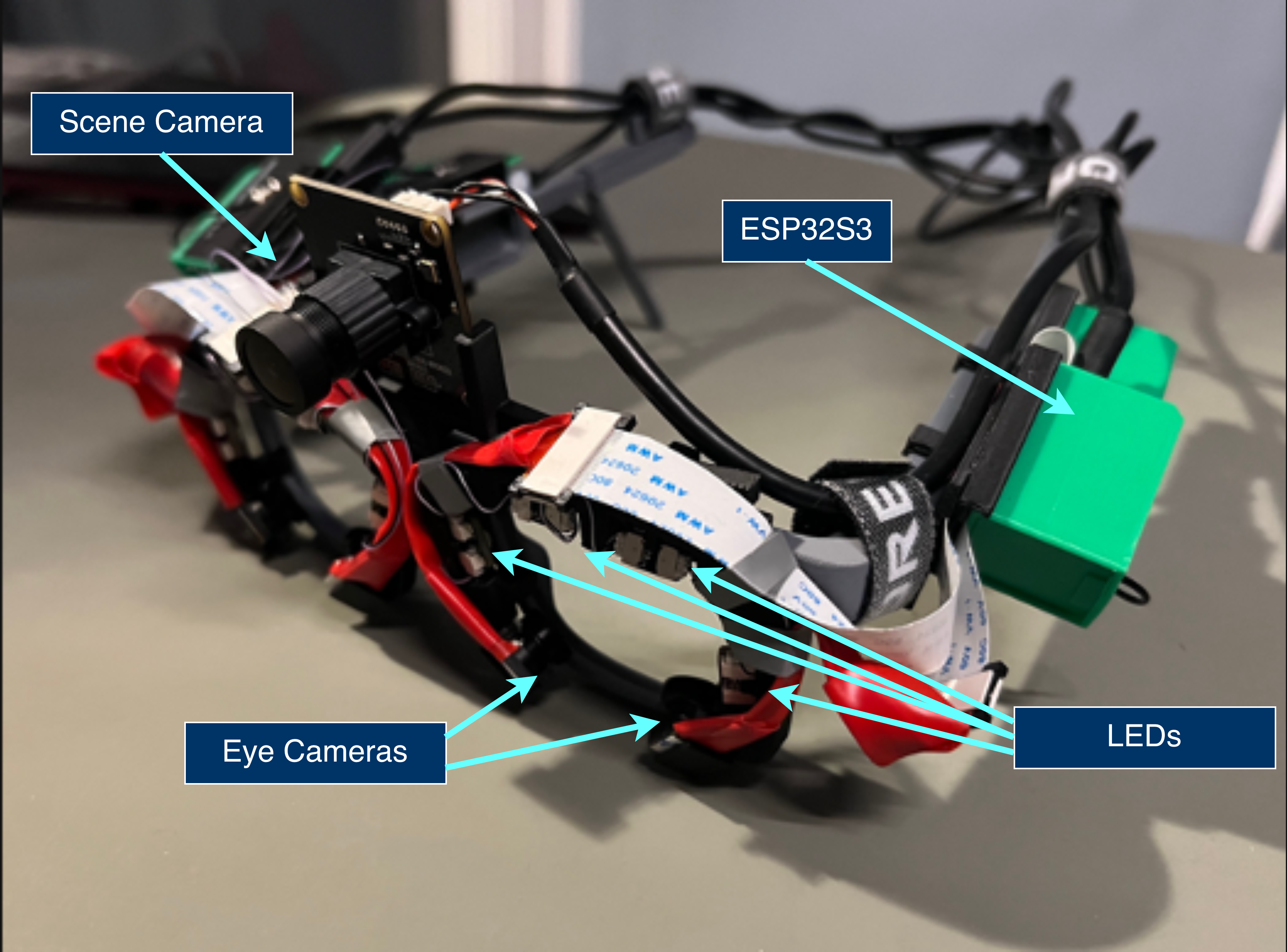}
        \caption{The full eye-tracker prototype with labeled components, using a stereo scene camera setup.}
        \label{fig:photo}
    \end{subfigure}
    \begin{subfigure}[t]{\linewidth}
        \centering
        \includegraphics[height=1in]{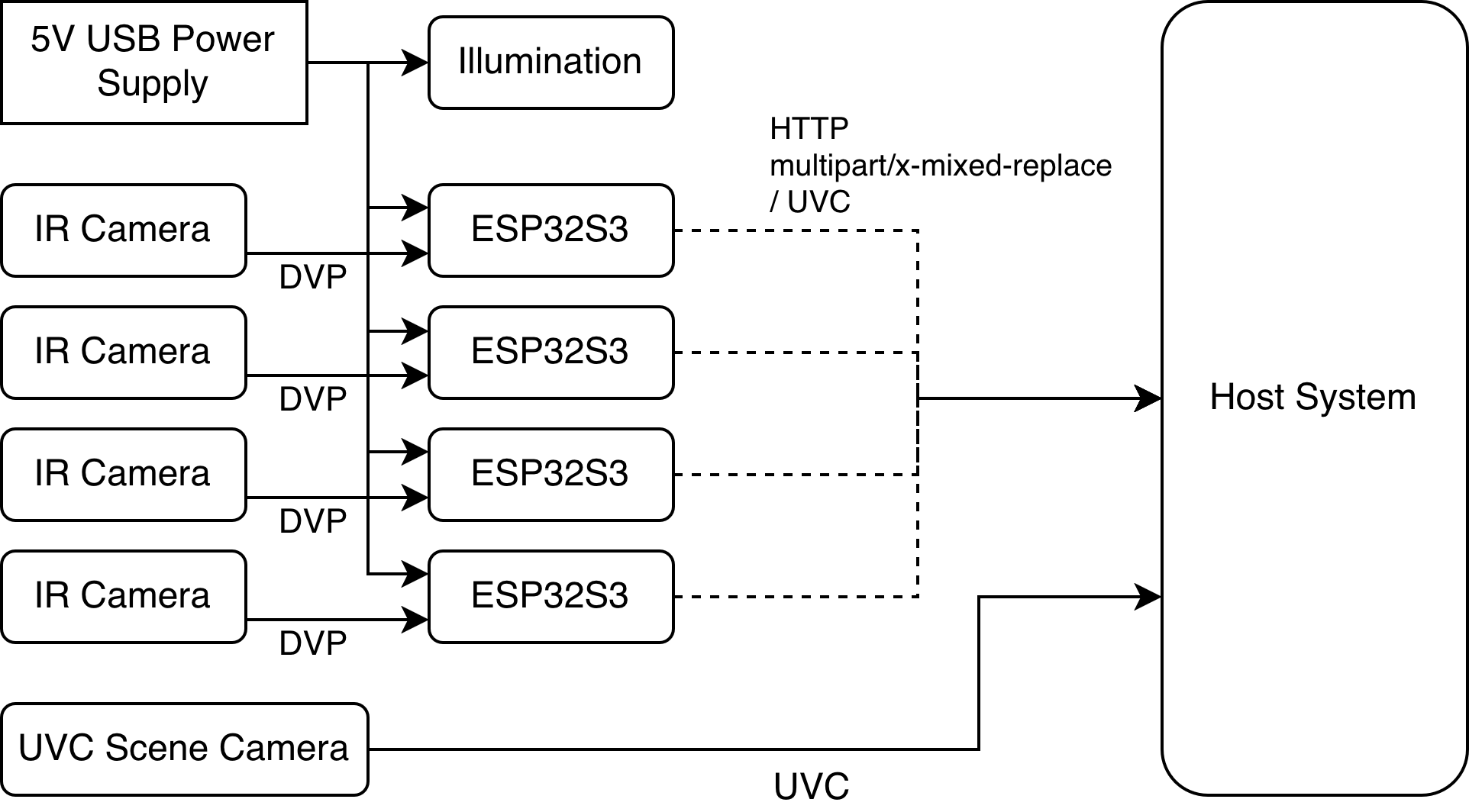}
        \caption{Schematic electronic hardware architecture of the eye-tracker.}
        \label{fig:arch}
    \end{subfigure}
    \begin{subfigure}[t]{\linewidth}
        \centering
        \includegraphics[height=1in]{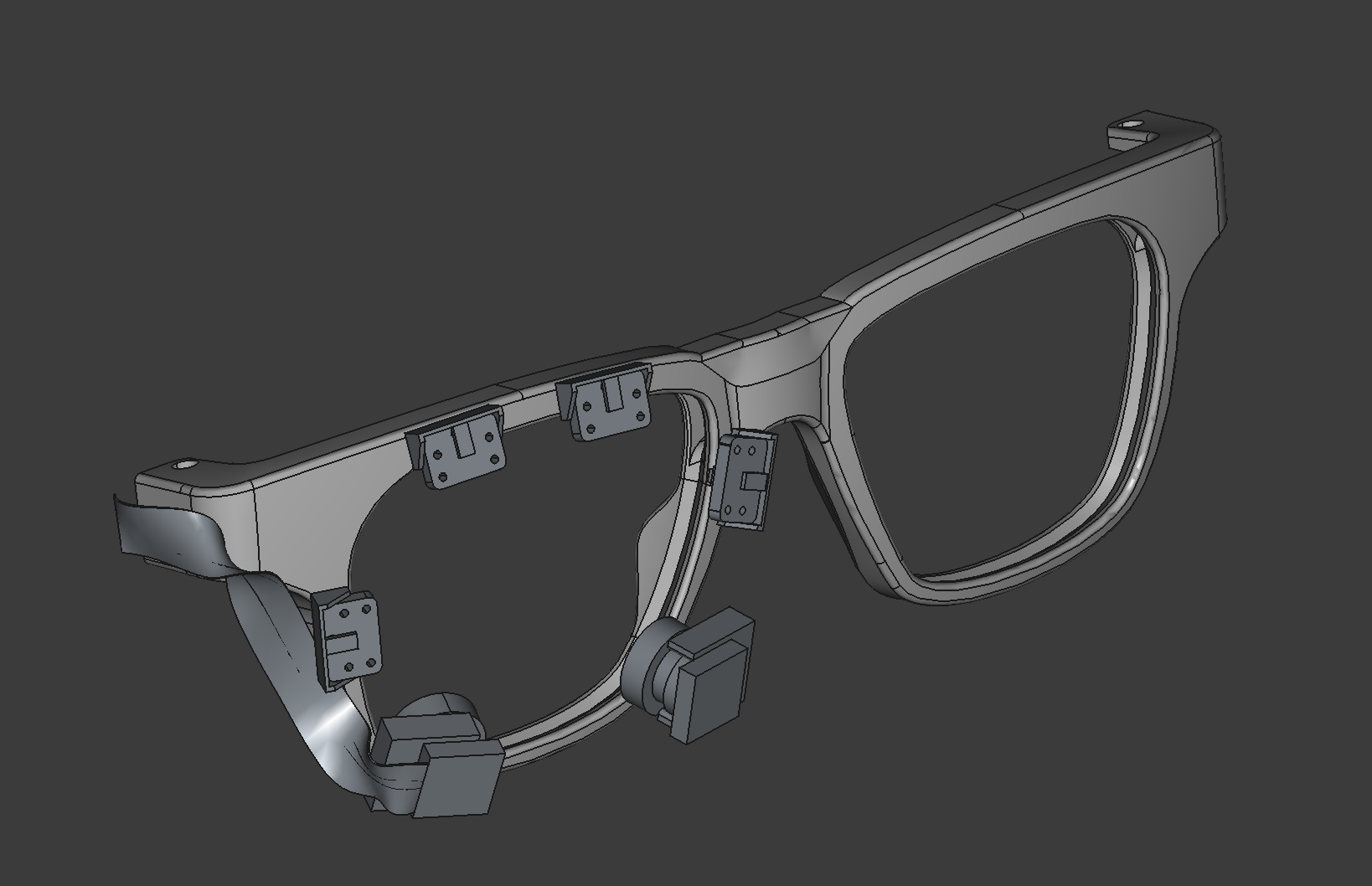}
        \caption{CAD rendering of the placement of the cameras and LEDs for one eye.}
    \end{subfigure}
    \begin{subfigure}[t]{\linewidth}
        \centering
        \includegraphics[height=1in]{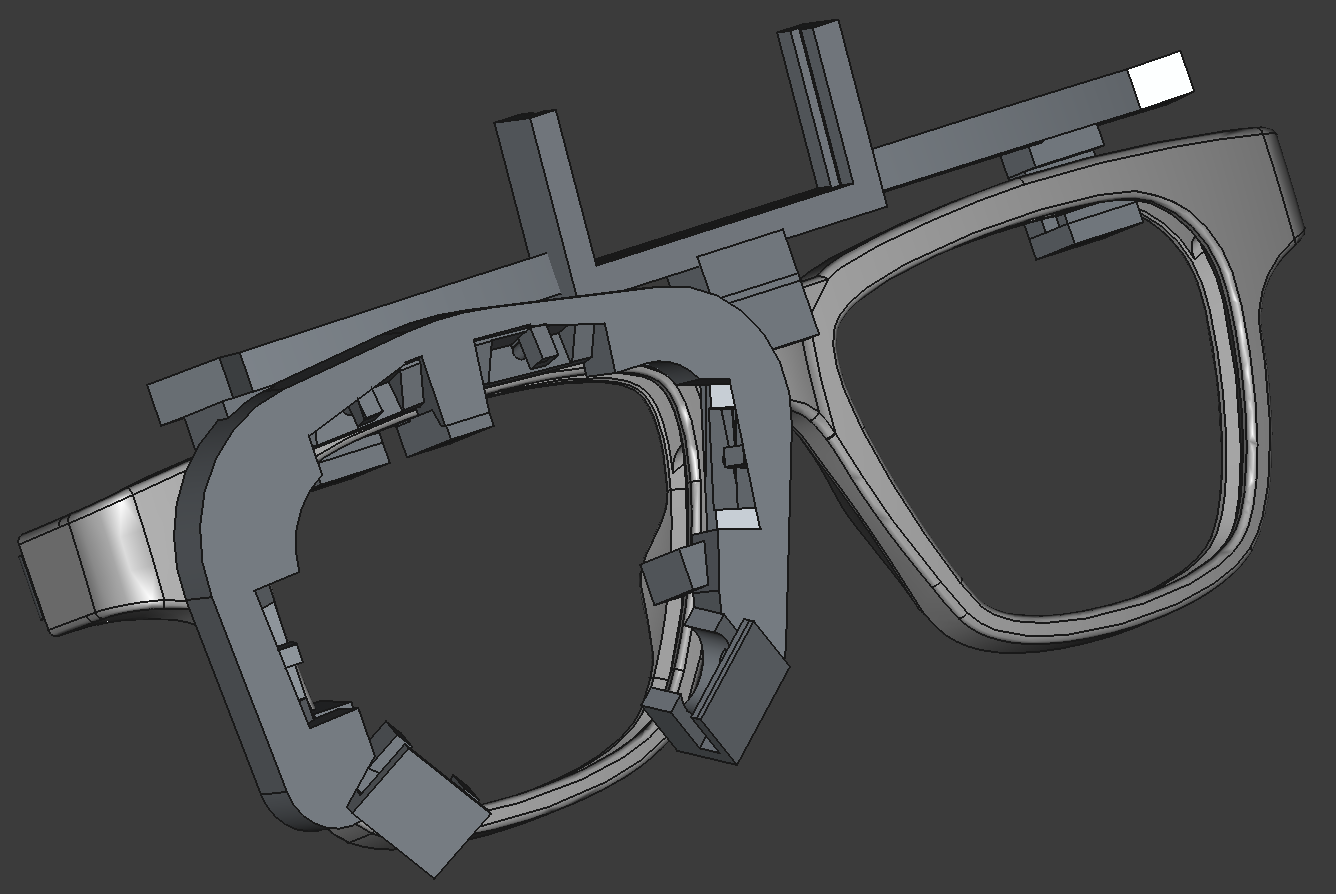}
        \caption{CAD design of the camera and LED mount, which clips into the printable frames.}
        \label{fig:cad}
    \end{subfigure}
    \caption{Overview of the eye tracker prototype.}
    \label{fig:overview}
\end{figure}

This Section details the architecture and individual hardware components used for the eye-tracker. An overview of the final prototype and its components is provided in Figure~\ref{fig:photo}.

\subsection{Architecture}

The developed system is a binocular eye-tracker, meaning that both eyes are tracked simultaneously. For each eye, a pair of wide-angle infrared cameras is used, enabling stereoscopic computer vision approaches. Four infrared LEDs per eye are used for illumination and as a source for glints. In addition, a scene camera setup is included, mounted in the center of the frames. The position of all cameras and LEDs are calibrated beforehand. Figure~\ref{fig:arch} schematically details the electronics of the system.

The setup is partly based on the EyeTrackVR~\cite{etvr} project. Reusing the firmware and hardware selection ensures that relevant components are reliable and safe for at least small-scale applications.

The system consists of rigid base frames without any electronics. All components are placed on additional mounts that can be clipped onto those frames. This allows for components to be easily swapped, thereby supporting a variety of frame sizes and designs.

With the exception of the scene camera, all cameras support wireless data transmission, allowing for easy portability as only a power supply is required for the frames and the data recording and processing can take place on any stationary system.


\subsection{Eye Cameras}

Due to a lack of affordable off-the-shelf UVC (USB Video Class) infrared cameras suitable for near-eye mounting, DVP (Digital Video Port) infrared wide-angle cameras are used in conjunction with one ESP32S3 microcontroller per camera, which is used to control and translate the DVP interface. The use of the DVP interface also allows for a wide selection of compatible cameras, allowing for further hardware comparison. In our case, an OV2640 sensor with a wide-angle 160 degree optic and infrared-compatible wavelength filter of 850 nm is used. The cameras need to be manually focused to the eye region and calibration-relevant distance during assembly.

The firmware for the ESP32S3 is that used by EyeTrackVR~\cite{etvr} with minor adaptations. It supports wireless or wired (UVC) transmission, with the latter being inherently more robust at the cost of portability.

\begin{figure}
    \centering
    \includegraphics[height=1.5in]{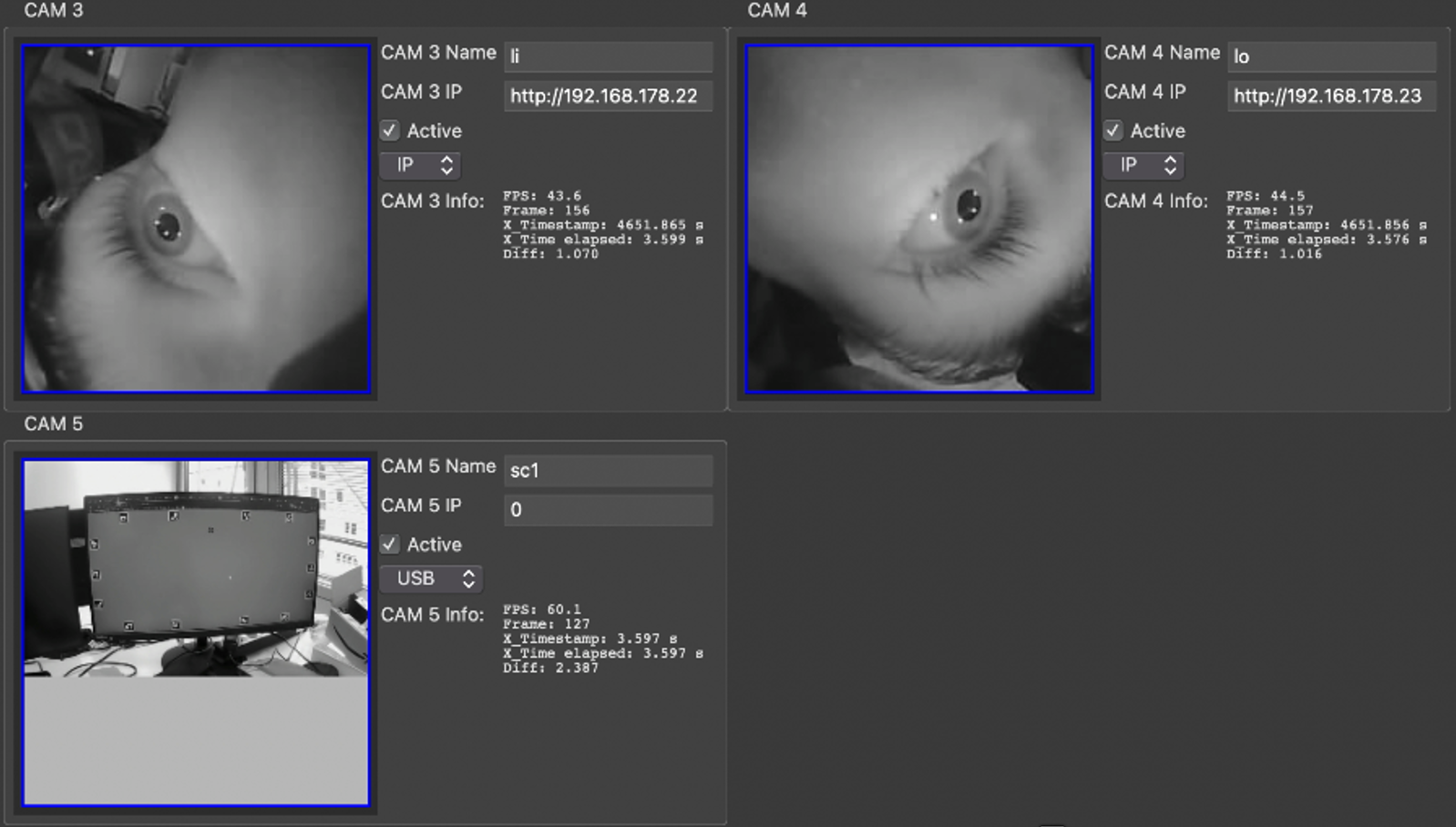}
    \caption{Sample recordings of a cameras pair and the scene camera in a typical indoor environment, using a custom GUI. Glints are visible as four bright spots on the cornea.}
    \label{fig:recordings}
\end{figure}

The achievable frame rate is approximately 45 FPS, precise frame timing is transmitted with every frame. The resolution is 240x240 pixels, which is sufficient for detecting the pupil outline and glints. Figure~\ref{fig:recordings} shows sample recordings of the cameras for one eye. The cameras are oriented at an angle and are not co-planar, which can be resolved via calibration as discussed in Section~\ref{sec:calibration}.

Sending streaming requests to all cameras simultaneously ensures a good synchronicity between all recordings. Alternatively, a GPIO pin can be used as a trigger to start the recordings. Any drift of the frame timings is corrected by the timing information of the frames, based on the internal clocks of the microcontrollers. However, due to the lack of a dedicated trigger channel, which is only found on larger and more expensive hardware, no sub-frame-perfect synchronization can be ensured, so frames can be offset relative to each other as frame drift accumulates. Some degree of motion blur can be present in the recordings as well, which can impact the robustness of some detection algorithms. Additionally, wireless transmission might lead to frame drops if the connection is slow or unstable.


\subsection{Scene Camera}

A scene camera is included to support gaze mapping and enable future extensions such as egocentric interaction analysis. Important are a small form-factor and reasonably high resolution and frame rate. For our prototype, we used a cheap, triggerable off-the-shelf UVC camera. A wireless solution similar to the eye-cameras, while technically possible, was ruled out due to limitations in resolution, which needs to be higher for the scene camera, and the availability of suitable UVC alternatives.

\subsection{Illumination}

A strand of four Infrared LEDs per eye, distributed around the frame, is used for illumination. The mounts are compatible with the PCB designs from the EyeTrackVR\cite{etvr} project. Eye-Safety was already evaluated as part of that project as well, which makes these components a preferred choice for the illumination.

\subsection{Mount}

The system is designed as a clip-on-solution for existing frames, which allows for easier adaptability to different frame sizes and shapes without having to remount all components individually. Notably, the cameras and LEDs are all mounted on a single carrier clip per eye, and similarly the ESP32S3s and control electronics are mounted on a base plate which can be attached to the hinges.
The frames feature working non-rigid hinges to avoid flexing of the components relative to each other, which would lead to the extrinsic calibration becoming invalid. This comes at the cost of a higher proneness to slippage, which can be mitigated by using a head strap or filtered by employing more sophisticated tracking algorithms.
All components are 3D-printable and can be mounted without additional components or specialized tooling. Figure~\ref{fig:cad} shows a CAD rendering of the mount and system components.

Due to the eye cameras being placed on the outside of the frames, the use in conjunction with prescription lenses or glasses leads to distortions and reflections and needs to be further evaluated.


\section{Eye Tracker Calibration} \label{sec:calibration}

Intrinsics and Extrinsic parameters of all cameras and the LEDs need to be known for stereo eye-tracking algorithms. This Section outlines the calibration procedure used to achieve this. A full system calibration can be completed in approximately 10 minutes once the setup is assembled.

\subsection{Eye Camera Calibration}

Cameras are calibrated in a pairwise fashion using a typical checkerboard-based calibration procedure, which simultaneously calibrates the extrinsic and intrinsic parameters. Calibration is performed in multiple stages to estimate intrinsic parameters, extrinsic camera poses, and LED positions in a shared coordinate system. The following steps are required in order to calibrate the eye cameras:

\begin{enumerate}
    \item Pairwise calibration of the left and right camera pairs using a smaller pattern, which is achieved by using standard libraries like OpenCV.
    \item Calibration of the left and right outermost cameras using a larger pattern while mounted on the frames. The focus range of the cameras must be adjusted beforehand to support both distances without refocusing.
\end{enumerate}

Each pairwise calibration requires recording multiple images of the calibration pattern in numerous different poses, which is aided by a custom software for capturing calibration patterns and performing the calibration.

\begin{figure}[t!]
    \centering
    \begin{subfigure}[t]{\linewidth}
        \centering
        \includegraphics[height=1in]{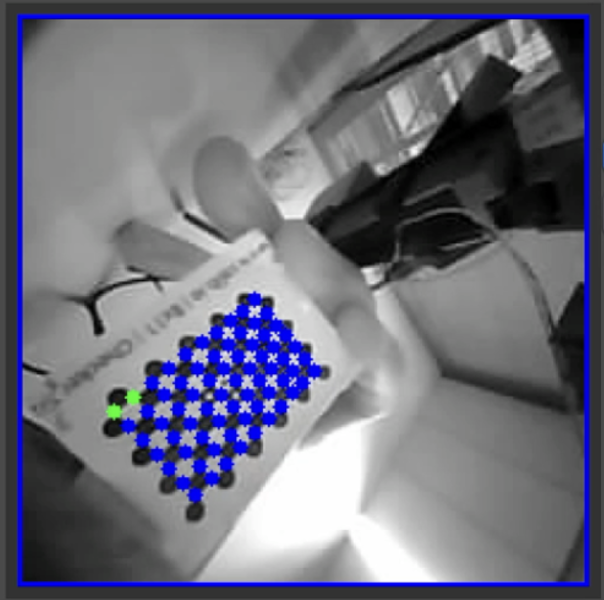}
        \caption{An eye camera recording of the calibration pattern.}
        \label{fig:calibration_pat}
    \end{subfigure}
    \begin{subfigure}[t]{\linewidth}
        \centering
        \includegraphics[height=1.2in]{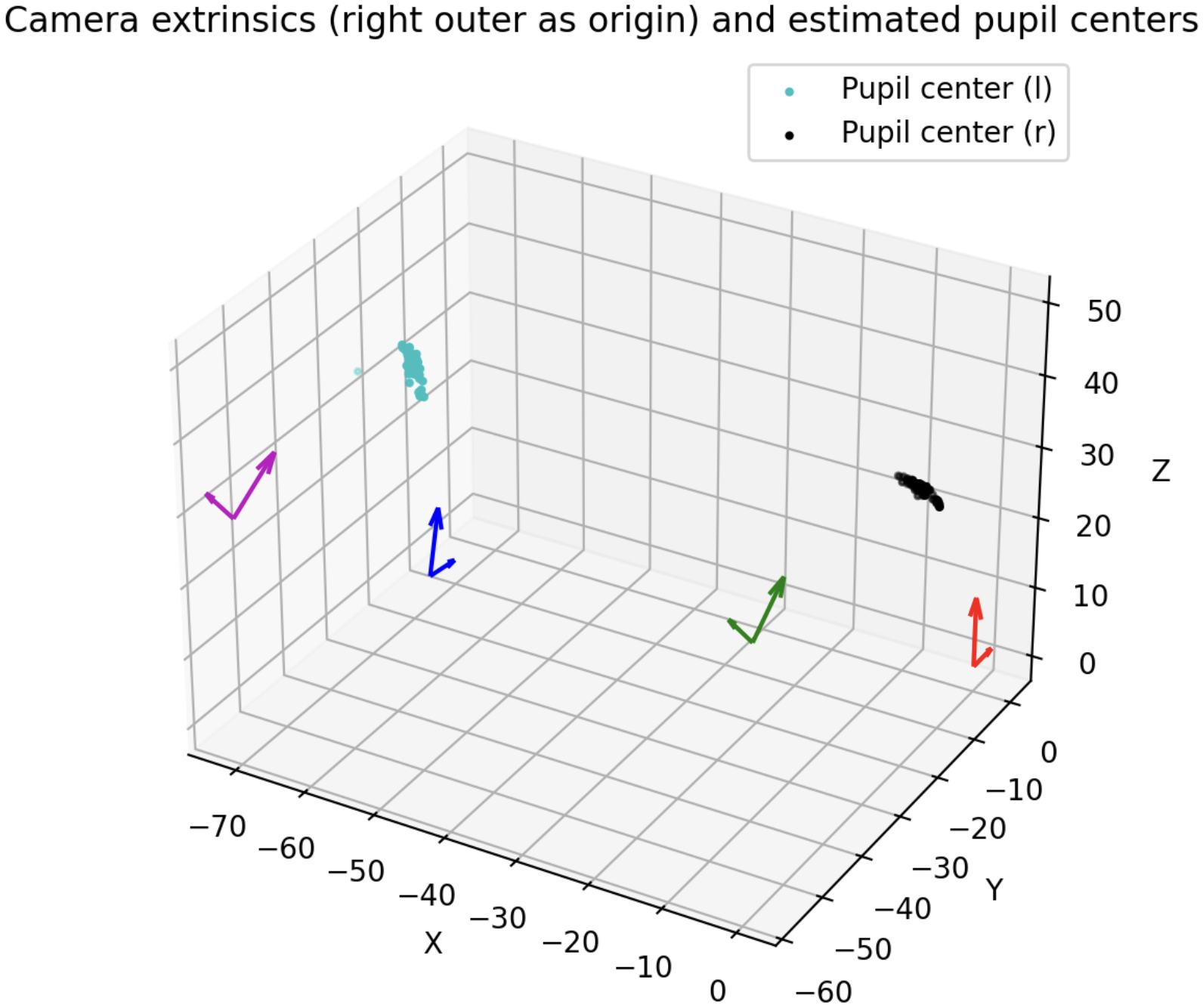}
        \caption{Calibrated camera positions and orientations with triangulated pupil center positions. The larger arrows indicate the camera viewing direction, while the smaller arrows mark the bottom of the image frame (the respective Y-axis). Scattered are the estimated pupil positions for a sample recording, based on a stereo reconstruction algorithm.}
        \label{fig:calibration_eye}
    \end{subfigure}
    \begin{subfigure}[t]{\linewidth}
        \centering
        \includegraphics[height=1in]{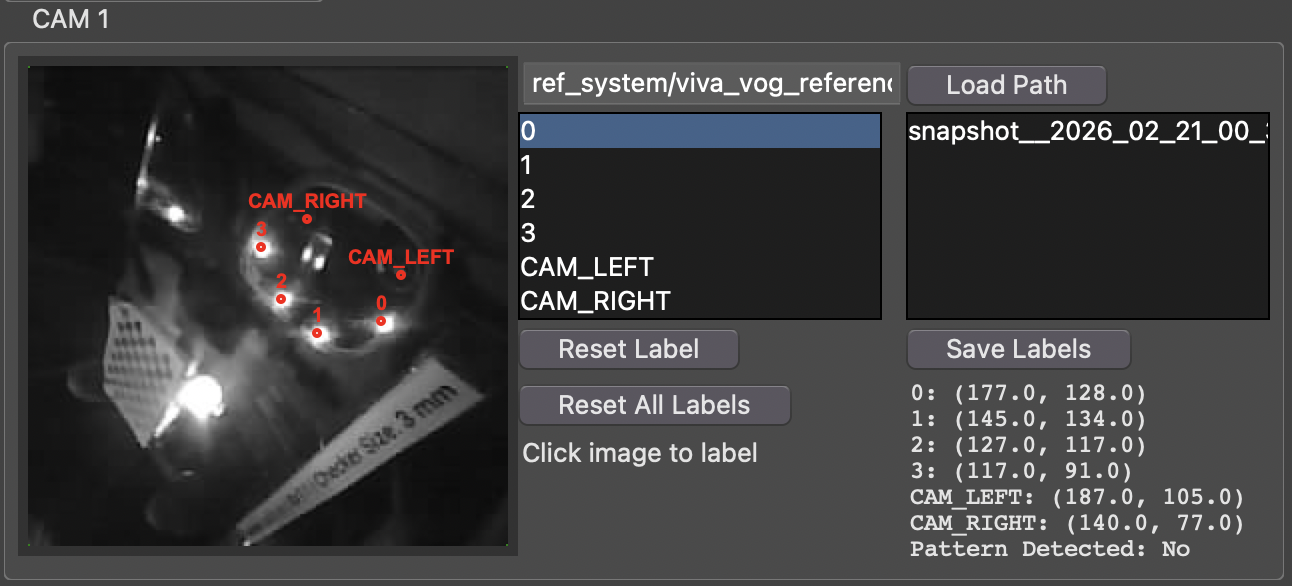}
        \caption{Labeling tool for the LED and scene camera positions.}
        \label{fig:calibration_led_label}
    \end{subfigure}
    \begin{subfigure}[t]{\linewidth}
        \centering
        \includegraphics[height=1in]{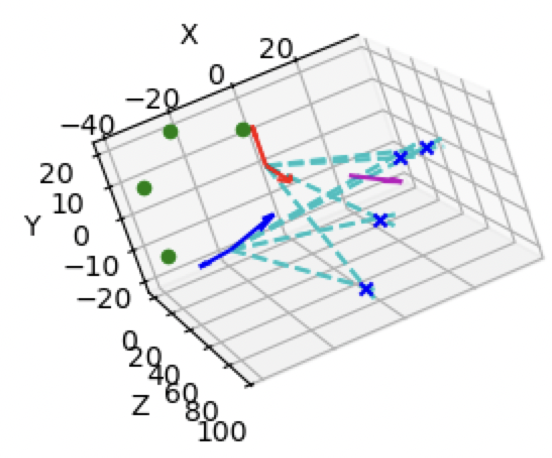}
        \caption{LED calibration with green dots corresponding to the calibrated LED positions above the cameras and blue crosses as their mirrored counterparts. The purple arrow indicates the surface normal of the mirror.}
        \label{fig:calibration_led}
    \end{subfigure}
    \caption{Visualization of some of the calibration steps.}
    \label{fig:calibration}
\end{figure}

Figure~\ref{fig:calibration_eye} shows the result of a full calibration of the eye cameras. A sample calibration capture is shown in Figure~\ref{fig:calibration_pat}.

\subsection{LED Position Calibration} \label{sec:ledcalib}

For calibrating the LED positions, a procedure similar to \cite{shihNovelApproach3D2004} is used per eye. A mirror with a mounted checkerboard pattern is positioned in a way such that all LED reflections are visible from both eye cameras. The orientation of the mirror is deducted from the position of the checkerboard pattern, and the reflected light spots are labeled. As the extrinsic parameters of both cameras are already known from the previous calibration, the location of the LEDs in the world coordinate system can be derived by triangulation.

Instead of using a pattern or markers, the cameras themselves can be labeled to reconstruct the pose of the mirror.


The LED placement is designed to support both symmetric and asymmetric glint configurations required by different model-based approaches.

Figure~\ref{fig:calibration_led_label} shows the labeling process, with the result visible in Figure~\ref{fig:calibration_led}.

\subsection{Scene Camera Calibration}

The calibration of the pose of the scene camera relative to the eye cameras is not as straightforward, as the fields of view of those cameras don't overlap. It can be achieved with the following approach:

\begin{enumerate}
    \item Mono calibration of the scene camera intrinsic parameters using the known checkerboard-based approach.
    \item Calibration of capturing a mirror with a mounted calibration pattern and using a procedure similar to the one described in section \ref{sec:ledcalib}.
        \begin{enumerate}
            \item First, the mirror is captured by two eye cameras and a marker on the back of the scene camera is labeled (corresponding to the center of the image sensor), yielding the position of the scene camera.
            \item Then, the same is done for the scene camera, instead labeling the positions of the eye cameras, which additionally gives the orientation of the scene camera.
        \end{enumerate}
\end{enumerate}

Finally, all all calibrated systems are transformed into a single world coordinate system, which is achieved by concatenating the gathered pairwise transformations with respect to a defined camera (e.g. the scene camera), which serves as the basis of the world coordinate system.

\section{Discussion}

\subsection{Supported Algorithmic Paradigms}

The presented hardware setup supports a wide range of eye-tracking algorithmic paradigms that are difficult or impossible to realize with webcam-based or monocular wearable systems. In particular, the system enables the development and evaluation of pupil detection and segmentation methods, including learning-based approaches such as EllSeg~\cite{kothariEllSegEllipseSegmentation2021}, classical methods such as PuRe \cite{santiniPuReRobustPupil2018} and contour-based gaze-estimation techniques~\cite{swirskiFullyautomaticTemporalApproach2013}. Beyond monocular processing, the use of two stereo eye camera pairs allows for stereo triangulation of pupil centers and related ocular features, supporting geometric gaze estimation methods that exploit binocular constraints~\cite{wang20133D}. In addition, the integrated infrared illumination enables glint-based gaze estimation approaches that rely on corneal reflections as stable reference points~\cite{guestrin2006General,barsingerhornDevelopmentValidationHighspeed2018}. The system further supports binocular recordings across both eyes, enabling the exploration of vergence-based depth estimation and focus-related measures, which are increasingly relevant for immersive and near-eye applications.

By supporting these paradigms within a single, unified hardware configuration, the platform allows researchers to compare and combine multiple approaches under consistent sensing conditions, facilitating hybrid methods that integrate geometric, photometric, and learning-based cues.

\subsection{Trade-offs and Limitations}

The focus on affordability and accessibility introduces several trade-offs that impact tracking performance. The use of low-cost DVP cameras and ESP32S3 controllers limits the achievable frame rate to approximately 45 FPS and constrains image resolution, which can reduce the robustness of pupil and glint detection under rapid eye movements. Furthermore, the absence of a full hardware trigger channel prevents sub-frame-perfect synchronization between cameras, leading to small temporal offsets that may affect stereo-based methods, particularly during fast motion.

Wireless data transmission, while improving portability, can result in frame drops under unstable network conditions, so the use of the serial interface is recommended if possible. The reliance on manually focused wide-angle optics introduces additional variability during assembly. The mechanical design, which prioritizes modularity and adaptability, also introduces a trade-off between calibration stability and user comfort. Although non-rigid hinges reduce flexing, which can otherwise invalidate the extrinsic calibration, they increase susceptibility to slippage, which can necessitate algorithmic compensation or filtering. Finally, the form factor and component selection lead to a larger physical footprint compared to highly integrated commercial systems. While this does not significantly occlude the user’s field of view, it limits applicability for long-term or everyday use and reinforces the platform’s role as a research and development tool rather than a consumer-grade device.

\subsection{Implications for Research}

Despite these limitations, the proposed platform represents an important step toward democratizing eye-tracking research by lowering barriers to entry for advanced algorithm development. By relying exclusively on off-the-shelf components and 3D-printable mounts, the system enables researchers to reproduce, modify, and extend the hardware without specialized manufacturing resources, supporting open and transparent experimentation. The ability to support stereo, glint-based, and binocular paradigms within a single wearable system enables systematic benchmarking and comparative evaluation of eye-tracking methods under consistent conditions, an aspect that is often missing in prior work due to hardware constraints. This, in turn, facilitates more principled investigation of algorithmic trade-offs and encourages the development of hybrid approaches that combine multiple sources of information. More broadly, the platform shifts the focus of research effort away from rebuilding bespoke hardware toward advancing algorithms, calibration methods, and theoretical models. By making the complete hardware design openly available, this work contributes to improved reproducibility and fosters a shared experimental foundation for future eye-tracking research.

\subsection{Privacy and Societal Impact Statement}

By using a self-contained, researcher-controlled eye-tracking system, the risk of unintended data sharing or misuse by third-party vendors is reduced. This is because raw eye-tracking data remain fully under the control of the researchers and study participants. This supports privacy-preserving research practices by enabling local data processing, transparent data handling, and informed consent without reliance on proprietary platforms or external cloud services.

\section{Conclusion}

We introduced a novel hardware setup suitable for the development and evaluation of various paradigms of eye-tracking algorithms. While its utility remains limited for end-user and high-end performance purposes, its focus on affordable and accessible hardware contributes to the democratization of fundamental eye-tracking algorithm research. All hardware designs and documentation are made openly available. 

\begin{acks}
The project is supported by the Chips Joint Undertaking (Chips JU) and its members, including top-up funding by Denmark, Germany, Netherlands, Sweden, under grant agreement No. 101139942.
\end{acks}

\bibliographystyle{ACM-Reference-Format}
\bibliography{literature}

\end{document}